\documentclass[journal]{IEEEtran}

\usepackage{blindtext}
\usepackage{graphicx}
\usepackage{amsmath}
\usepackage[utf8]{inputenc}
\usepackage{hyperref}
\usepackage{color}
\usepackage{cleveref}
\usepackage{setspace}
\usepackage{tabularx}
\usepackage{etoolbox}
\usepackage{booktabs}
\usepackage{float}
\usepackage{placeins}
\usepackage{ragged2e}

\newcommand\ppl{\mathit{ppl}}
\newcommand\mppl{\mathit{mppl}}
\renewcommand{\vec}[1]{\boldsymbol{#1}}

\newcolumntype{R}[1]{>{\RaggedLeft\arraybackslash}p{#1}}

\newcommand{\Revision}[1]{#1}
\newcommand{\RevisionOnly}[1]{}

\begin{document}
\title{The KIT Motion-Language Dataset}

\author{Matthias~Plappert,
        Christian~Mandery,
        and~Tamim~Asfour \\[5mm]
        \small{Article originally appeared in \emph{Big Data}, Vol. 4, No. 4, pp. 236--252, 2016. \\ Final publication is available from Mary Ann Liebert, Inc., publishers \url{http://dx.doi.org/10.1089/big.2016.0028}.}
\thanks{All authors are with the High Performance Humanoid Technologies (H\textsuperscript{2}T) lab, Institute for Anthropomatics and Robotics (IAR), Karlsruhe Institute of Technology (KIT), Adenauerring 2, 76131 Karlsruhe, Germany. E-mail: matthiasplappert@me.com, \{mandery, asfour\}@kit.edu.}}
\markboth{Plappert \MakeLowercase{\textit{et al.}}: The KIT Motion-Language Dataset}{}

\maketitle

\begin{abstract}
  Linking human motion and natural language is of great interest for the generation of semantic representations of human activities as well as for the generation of robot activities based on natural language input.
  However, while there have been years of research in this area, no standardized and openly available dataset exists to support the development and evaluation of such systems.
  We therefore propose the \emph{KIT Motion-Language Dataset}, which is large, open, and extensible.
  We aggregate data from multiple motion capture databases and include them in our dataset using a unified representation that is independent of the capture system or marker set, making it easy to work with the data regardless of its origin.
  To obtain motion annotations in natural language, we apply a crowd-sourcing approach and a web-based tool that was specifically build for this purpose, the \emph{Motion Annotation Tool}.
  We thoroughly document the annotation process itself and discuss gamification methods that we used to keep annotators motivated.
  We further propose a novel method, perplexity-based selection, which systematically selects motions for further annotation that are either under-represented in our dataset or that have erroneous annotations.
  We show that our method mitigates the two aforementioned problems and ensures a systematic annotation process.
  We provide an in-depth analysis of the structure and contents of our resulting dataset, which, as of \Revision{October 10, 2016, contains 3911 motions with a total duration of 11.23~hours and 6278 annotations in natural language that contain 52\,903~words}.
  We believe this makes our dataset an excellent choice that enables more transparent and comparable research in this important area.
\end{abstract}

\section{Introduction}
\label{sec:intro}
Human motion plays an important role in many fields, including sports, medicine, entertainment, computer graphics, and robotics.
Today, a wide variety of commercial motion capture systems exist that can be used to record vast amounts of motion data.
In robotics, observation of human subjects helps to improve understanding of how humans succeed in challenging environments and perform complicated tasks.
Great research effort has been put into the recording, processing, storage, and transfer of human motion.
The collected data offers a promising way towards the intuitive programming of robot systems with humanoid embodiments.
However, observing only the motion of a human teacher is not sufficient as a teacher very often includes additional or corrective instructions to the student using natural language.
In other words, a teacher-student interaction and therefore such \emph{programming by demonstration} concept~\cite{DBLP:reference/robo/BillardCDS08} is inherently multi-modal.
Natural language also offers an intuitive way of describing complex motions and parametrizations thereof.
Take, for example, the very simple sentence ``A person wipes the table with their right hand 5 times''.
This single sentence encodes a rather complex motion (``wiping'') and even parametrizes the execution of the motion (``right hand'' and ``5 times'').
We therefore argue that the combination of motion and natural language plays a crucial role in achieving rich, multi-modal human-robot interactions.

\begin{figure}[t]
  \centering
  \includegraphics[width=0.7\columnwidth]{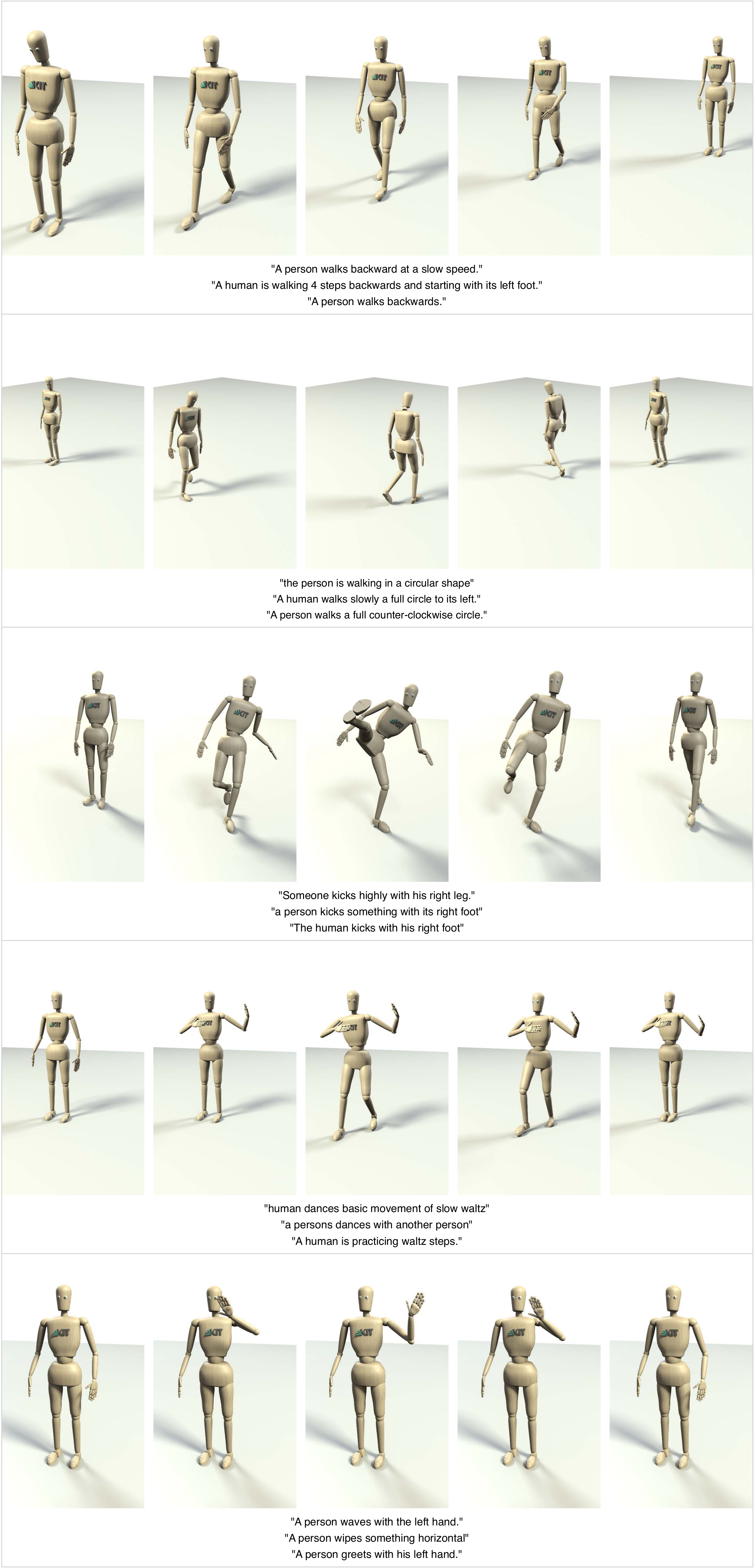}
  \caption{Five exemplary motions and their respective annotations in natural language from our dataset.}
  \label{fig:dataset}
\end{figure}

Besides these long-term goals of building truly collaborative, easy and intuitive to program robots, the combination of motion and natural language also offers immediate applications. A common problem when dealing with motion is the retrieval of matching entries from a large motion databases.
In such a database, each entry is typically annotated with one or more labels, e.g. ``wiping'' and ``right hand''.
However, it is often difficult to select all appropriate labels during the manual annotation of a new motion.
Even more so, during the retrieval phase, the user must know exactly what he or she is looking for in order to specify the appropriate labels.
In this case, natural language offers a much richer and intuitive way of describing motions.
For new motions, rich descriptions could be generated automatically.
For retrieval queries, these rich descriptions could then be used to perform full-text search and also provide means of selecting appropriate entries from a list.
When using a generative system, a query in natural language could even be used to synthesize the requested motion.

While there has been active research to link human motion and natural language~\cite{DBLP:journals/adb/SugitaT05, DBLP:conf/humanoids/TakanoN08}, a publicly available dataset that combines human motion and natural language does not exist currently.
As a result, different datasets have been used by researchers, which makes it hard to compare results.
In addition, potentially interested researchers cannot easily contribute since they lack the necessary data.
To overcome these problems, we propose the \emph{KIT Motion-Language Dataset}, which combines human motion and descriptions thereof in natural language.
We also systematically and thoroughly describe our methods to acquire the data and the contents of the resulting dataset.
This includes a novel method we developed during data collection to select motions for further annotation in systematic fashion.
This method, which we refer to as \emph{perplexity-based} selection, ensures that motions that are either under-represented in our dataset or that have contradicting annotations are preferred for further annotation, mitigating the two aforementioned problems.
As such, our dataset serves as an excellent candidate for a benchmark dataset, which is an import step towards more transparent, comparable and accessible research in this area.

The rest of this paper is organized as follows:
In \Cref{sec:related-work}, we show that the need for such a unified dataset is real by covering existing research in this area and the different datasets used to conduct it.
Next, we discuss the methods we used to acquire the data.
We first describe both modalities of our dataset separately, starting with human motion in \Cref{sec:motion}.
In \Cref{sec:natural-language}, we describe the crowd-sourced acquisition of descriptions in natural language and present a novel method that ensures a systematic annotation process.
Both, motion and natural language, are finally combined into our dataset, which we present in detail in \Cref{sec:dataset}.
\Cref{sec:perplexity} discusses the results of our novel sampling approach that we applied during the crowd-sourced annotation process.
Finally, we summarize our work in \Cref{sec:conclusion} and provide an outlook on what we believe are important directions for future work in this area.

\section{Related Work}
\label{sec:related-work}
In the last years, several large-scale databases of human whole-body motion have been acquired using optical marker-based motion capture techniques.
The KIT Whole-Body Human Motion Database~\cite{DBLP:conf/icar/ManderyTDVA15, ManderyTerlemez2016}, see also,\footnote{\url{https://motion-database.humanoids.kit.edu/}} provides a rich corpus of human whole-body motion and contains freely available recordings of a wide range of motion types, such as locomotion, manipulation, loco-manipulation, gesticulation, and interaction.
The CMU Graphics Lab Motion Capture Database~\cite{DB_CMU} also provides an open dataset of whole-body motion, which covers an comparable selection of motion types and is well-established in several fields such as robotics and computer animation.
As these databases were used for the creation of our dataset, they are explained in more detail in \cref{sec:motion}.
HDM05~\cite{Mueller2007_HDM05} provides a dataset of 1457 human whole-body motion clips with a total run length of around 50 minutes, which have been created by segmenting a \Revision{continuous} motion sequence demonstrated by five different non-professional actors.
The Human Motion Database~\cite{Guerra2012_HumanMotionDatabase} provides five different datasets that have been acquired by using a systematic sampling methodology to select motions to be collected and additionally provides a survey of some existing motion databases in the cited article.
In the Edinburgh CGVU Interaction Database~\cite{DB_CGVUInteractionDatabase}, human whole-body motion for manipulation and object interaction tasks is captured using magnetic and RGB-D sensors.
The NUS Motion Capture Database~\cite{DB_NUS} contains whole-body motion capture recordings of eight different subjects for locomotion tasks and sports motions such as dance and martial arts.
The Human3.6M Dataset~\cite{Ionescu2014_Human36M} provides a large-scale dataset for the evaluation for human pose recognition methods, which contains complementary data from time-of-flight cameras and 3D laser scans of the human subjects.
Mandery~et~al.\cite{ManderyTerlemez2016} provide a more in-depth discussion of existing human motion databases regarding size, methodology, and available motion types, also including databases that have specialized on more specific types of motion instead of whole-body actions.

While many datasets of human motion exist, none of the above-mentioned databases of whole-body motion includes textual descriptions beyond simple tags or keywords and, to the best of our knowledge, there are no publicly available datasets that combine human motion and natural language.
However, there has been active research in this area, which uses datasets that combine natural language and motion.

Sugita~et~al.~\cite{DBLP:journals/adb/SugitaT05} studied the interaction between linguistic and behavioral processes.
In their work, the authors used a mobile robot in an environment with colored objects.
The goal of the robot was to point at, push, or hit the red, blue, or green object.
The robot commands were articulated in natural language.
The authors used a dataset that consists of $18$~sentences of very simple structure, e.g. ``point green'' or ``push red''.
These sentences were combined with the recording of $90$~sensory-motor sequences, which were obtained by remote-controlling the robot.
Ogata~et~al.~\cite{DBLP:conf/iros/OgataMTKO07} described a similar approach to combine motion and natural language.
The authors used the arm of a humanoid robot, which had $4$~degrees of freedom (DoF).
The task of the robot was to move its arms to one of four colored areas on a table, as instructed in natural language.
The authors collected the dataset for training their model by remotely operating the robot arm and annotating each motion with a simple instruction, e.g. ``move to red slowly''.
A total of $24$~sentences, which consist of $17$~distinct words, were used to annotate $8$~different movements of the arm of the robot.
In a later work, Ogata~et~al.~\cite{DBLP:conf/riiss/OgataO13} used a similar dataset that contains $48$~motions of a $2$~DoF robot and $102$~simple sentences, using a total of $16$~distinct words.
None of the aforementioned data is publicly available.

Takano~et~al. have studied a statistical model that combines human motion and natural language extensively over many years.
In an early work~\cite{DBLP:conf/humanoids/TakanoN08}, the authors used a dataset which consists of $10$~human whole-body motions.
The motion data was recorded using an optical marker-based motion capture system and converted to a kinematic model of the human body with $20$~DoF to obtain a joint angle representation.
Each motion was then annotated with a single sentence, resulting in a total of $10$~sentences, which consist of $15$~distinct words.
The same authors report~\cite{DBLP:conf/icra/TakanoN09} on a similar, but slightly larger dataset with a total of $23$~pairs of motions and annotations thereof, which consist of $24$~distinct words.
In recent publications~\cite{DBLP:conf/icra/TakanoN12, DBLP:journals/ijrr/TakanoN15, takano2016generating}, Takano~et~al. used a dataset with vastly increased size.
The larger dataset contains a total of $467$~motions, which were, again, recorded using an optical motion capture system.
As in their previous work, the authors represented the motion using a kinematic model of the human body, which now features $34$~DoF.
The motions used in this dataset also seem to be far more diverse than before.
While not quantified by the authors, the samples listed suggest that a wide variety of different motions like walking, dancing, object manipulation, playing tennis, climbing stairs, and more were used in this new dataset.
In contrast, previous work by the authors focused on a very narrow selection of motion, which were obtained from a baseball player.
All motions were annotated with a total of $764$~sentences, which consist of $241$~distinct words.
In all cases, the annotations were originally written in Japanese and translated to English by the authors.
The datasets used by the authors are not publicly available.

Recent work of Takano~\cite{takano2015learning} attempted to build a larger dataset by crowd-sourcing the annotation problem.
The motion data used by the author was recorded using $17$~wearable \emph{inertial measurement units} (IMUs).
Similar to previous work, the recorded data was converted using a kinematic model of the human body, again with $34$~DoF.
The annotation process was carried out using a simple and publicly available web tool.\footnote{\url{http://www.ynl.t.u-tokyo.ac.jp/~takano/}}
For the annotation process, the motion was visualized using a rendered video of the kinematic model.
This means that the perspective of the motion shown to the user was fixed and could not be adjusted by the user during annotation.
The author asked participants to simultaneously segment the motions by providing time stamps as well as a description for each segment using a single sentence in English.
The resulting dataset consists of a total of $621$~motions, all of which were performed by the same subject.
The motions were recorded during office work and lectures, which the subject held.
The author states that $419$~distinct words were used to describe the motions.
While the paper does not mention how many sentences were collected, the publicly available statistics from the annotation web tool list a total of $2504$~annotations by $19$~annotators at the time of this writing.
The crowd-sourced dataset is not publicly available.

The combination of human behavior and natural language has also been an active research topic in the computer vision community.
Kuehne~et~al.~\cite{DBLP:conf/iccv/KuehneJGPS11} proposed the \emph{HMDB}~dataset, which contains $7000$~video clips.
The clips originated from different data sources like YouTube, Google Video, and digitized movies.
All video clips were manually annotated with $51$~different action categories like ``smile'', ``talk'', ``climb stairs'', ``swing baseball bat'', and many more.
\Revision{Similarly, the YouTube-8M Dataset~\cite{abu2016youtube} provides the URLs to $8$~million YouTube videos as well as annotations using $4800$~different scene-level tags like ``basketball'', ``mountain biking'', or `` cooking''.}
However, \Revision{both datasets do} not contain descriptions in complete sentences.
\RevisionOnly{moved down}Chen~et~al.~\cite{DBLP:conf/acl/ChenD11} proposed the \emph{Microsoft Research Video Description Corpus}.
The authors used a crowd-sourcing approach to collect descriptions in natural language of short YouTube video clips depicting a variety of different day-to-day actions, e.g. cutting a cucumber.
The dataset contains $2089$ video clips that were annotated with $85\,550$ descriptions in English.
The descriptions were created by $688$~paid workers on Amazon Mechanical Turk.
More recently, Torabi~et~al.~\cite{DBLP:journals/corr/TorabiPLC15} proposed a dataset that uses the \emph{Descriptive Video Services}, which is available on most DVDs, to extract natural language descriptions of video clips.
The resulting dataset contains $84.6$~hours of annotated video from $92$~DVDs and a total of $55\,904$~sentences.
The same approach was used by Rohrbach~et~al.~\cite{DBLP:conf/cvpr/RohrbachRTS15} to collect a large dataset of video clips and descriptions thereof in natural language.
Their dataset contains $68\,000$~annotations for video snippets from $94$~HD movies.
While these datasets are useful to create annotations from video clips, they are less suitable for the training of generative models that produce the requested motion, which is of especially great interest in robotics.

\section{Human Motion}
\label{sec:motion}
In this section, we discuss the first modality of our dataset: human motion.
We briefly present the procedures for the acquisition of motion data and describe the unified representation of whole-body motion that is used in our dataset.

\subsection{Acquisition of Human Motion}
A wide variety of commercially available motion capture systems exist to record motion data.
Typically, one differentiates between optical (e.g. based on stereo video, depth information, or the tracking of markers)~\cite{DBLP:journals/cviu/MoeslundHK06} and non-optical systems (e.g. based on IMUs)~\cite{DBLP:conf/humanoids/MillerJKM04}.
At the time of this writing, our dataset only contains motion data that was recorded using optical marker-based systems.
In such systems, light in the infrared spectrum that is emitted by the cameras is reflected by markers attached to the subjects (and optionally objects) of interest and the reflection is recorded.
Since the position of each camera is known from an initial calibration procedure and, generally speaking, each marker is visible to multiple cameras, the system can compute the coordinates of each marker in Cartesian space using triangulation.

We use data from the \emph{KIT Whole-Body Human Motion Database}~\cite{DBLP:conf/icar/ManderyTDVA15}, which was captured using a sampling frequency of $100$~Hz.
For human subjects, a standardized marker set that consists of $56$~markers is used.
The location of each marker on the human body is precisely specified and was derived from anatomical landmarks~\cite{DBLP:conf/icar/ManderyTDVA15}.
Our dataset also contains motion capture data from the well-established \emph{CMU Graphics Lab Motion Capture Database}~\cite{DB_CMU}.
The CMU data uses a different marker set~\cite{CMU_markerset}, which consists of $41$~markers.
Motions were recorded using either a sampling frequency of $60$~Hz or $120$~Hz.
In all cases, the captured motion data is stored in C3D~files,\footnote{\url{https://www.c3d.org/}} which is an industry standard.
These files contain the Cartesian coordinates of each marker for each time frame as well as additional metadata (e.g. marker names).

\begin{figure}[h]
  \centering
  \includegraphics[width=\columnwidth]{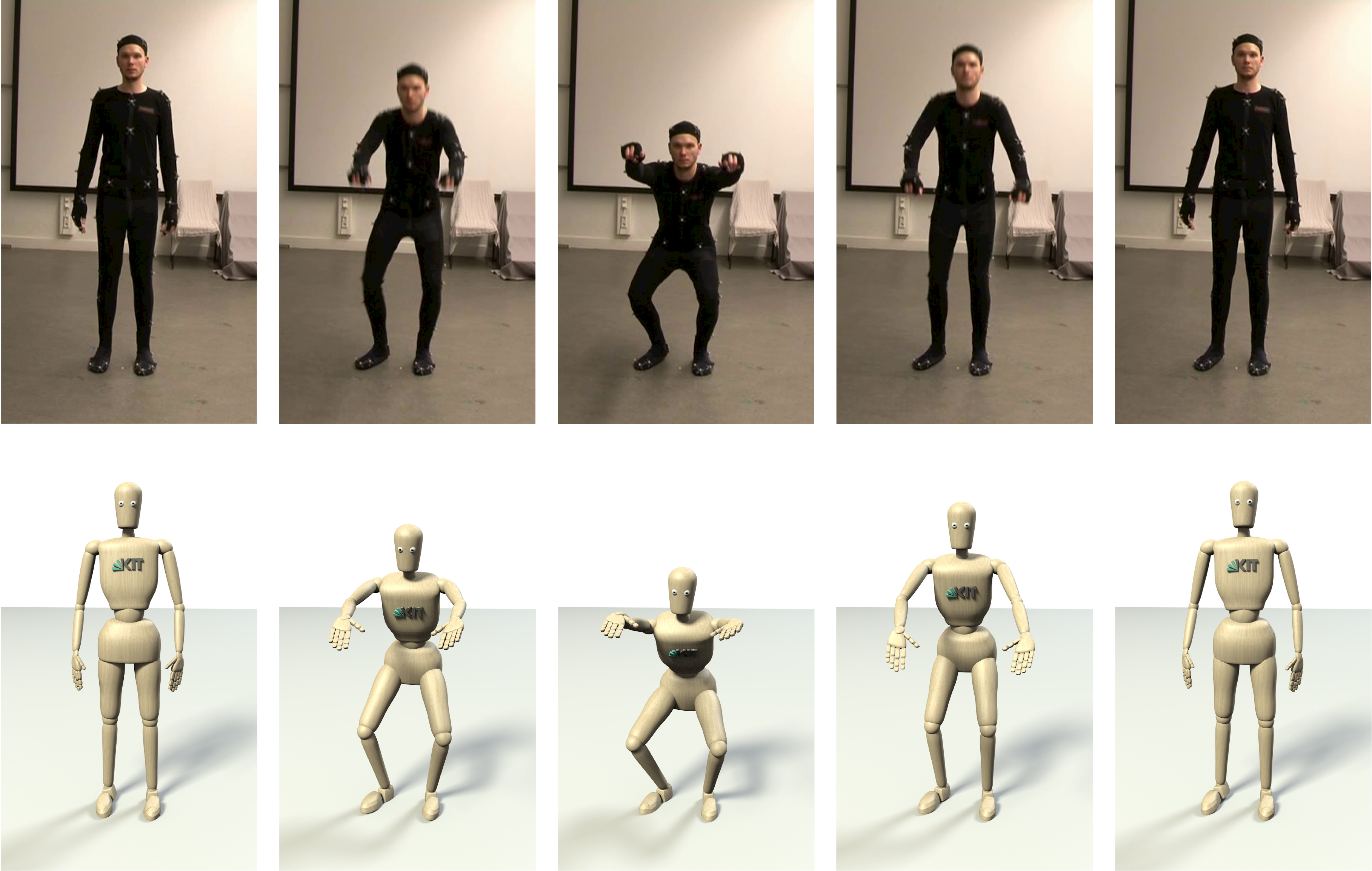}
  \caption{The conversion process from marker-based motion capture data (top row) to the Master Motor Map (MMM) representation (bottom row).}
  \label{fig:motion2mmm}
\end{figure}

\subsection{Representation of Human Motion}
While the previously described data already represents human motion, it does still depend on the given recording setup and especially the placement of motion capture markers.
Therefore, our goal is to achieve a unified representation that is independent of the motion capture setup and also inter-subject variations like the height.
This is essential to our goal of creating an open and extensible dataset since it allows us to convert motion data from multiple data sources that use different marker sets or even completely different capture systems to the same representation.
Luckily, this problem has already been solved in the past.

\begin{figure}[h]
  \centering
  \includegraphics[width=0.8\columnwidth]{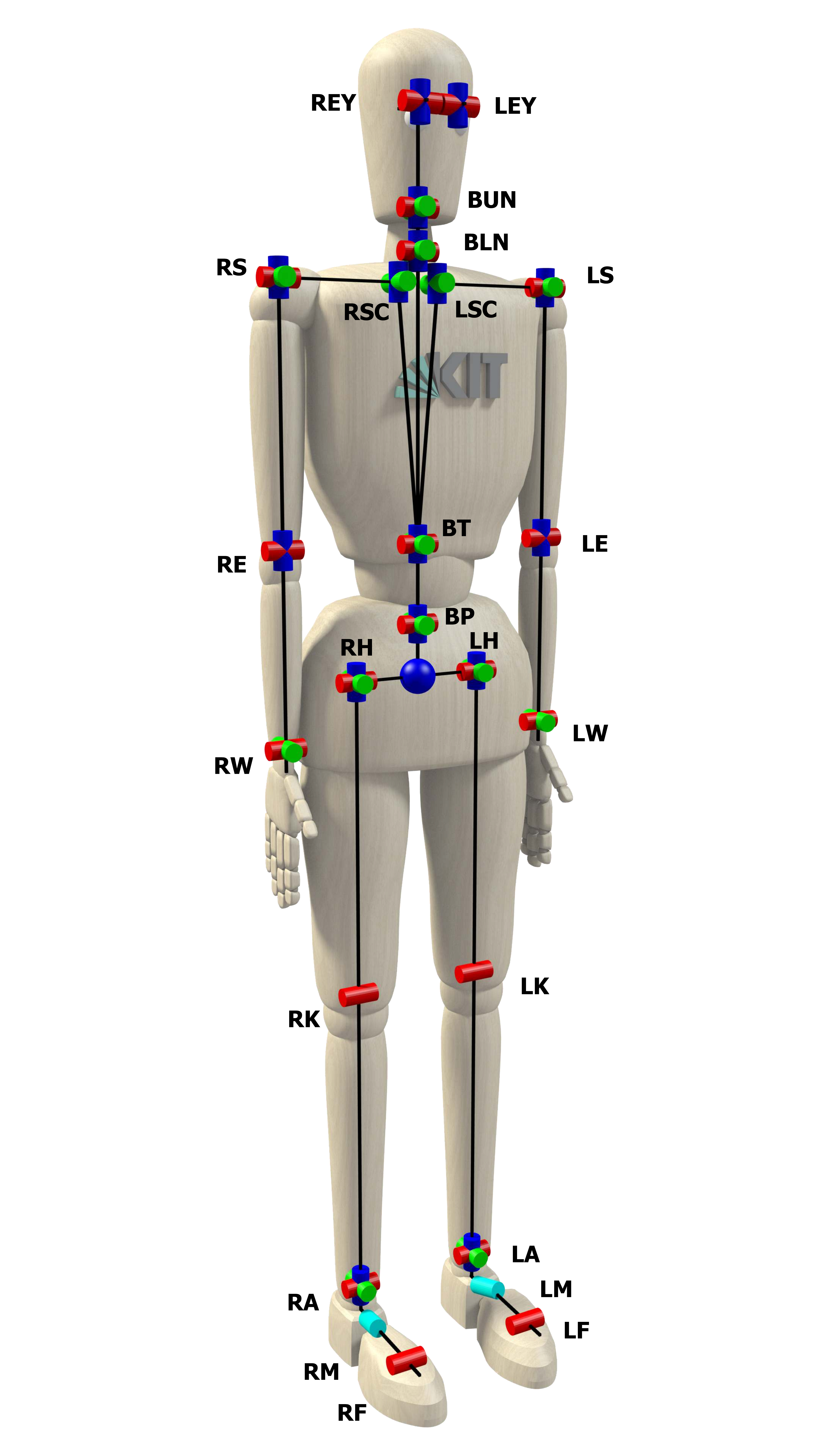}
  \caption{\RevisionOnly{added figure}The kinematics of the MMM reference model.}
  \label{fig:mmm}
\end{figure}

We use the \emph{Master Motor Map}~(MMM) framework~\cite{ManderyTerlemez2016, DBLP:conf/humanoids/TerlemezUMDVA14, DBLP:conf/icra/AzadAD07} for such a unified representation of human motion.
\Revision{The MMM framework provides converters to map the raw data recorded from various motion capture systems to a standardized reference model of the human body.
For a marker-based motion capture approach, which is, as previously mentioned, the motion acquisition technique used in this dataset, this is achieved by placing virtual markers on the reference model and computing the inverse kinematics such that the mean squared distance between corresponding physical and virtual markers is minimized.
To adopt the conversion process to data sources that use a different marker set, it is sufficient to update the marker placement on the 3D model accordingly.
For entirely different capture techniques that may be used in the future, the design of MMM framework allows for the implementation of additional converters as necessary.}
\Cref{fig:motion2mmm} illustrates the conversion process.
The kinematics of the MMM model are derived from well-established work on the human biomechanics~\cite{winter2009biomechanics} and uses $104$~DoF: $6$ DoF cover the model pose, $23$ DoF are assigned to each hand, and the remaining $52$ DoF are distributed on arms, legs, head, eyes, and torso.
While the high number of DoF is important to enable the representation of a wide range of human motion, only a subset of the DoF must be specified in order to use the model.
In this work we only use $50$~DoF ($6$~DoF for the model pose and $44$~DoF are distributed on arms, legs, head, and torso) since the individual fingers and eyes are not tracked.
\Revision{\Cref{fig:mmm} depicts the kinematics of the MMM reference model.}
The model also defines dynamic properties for each segment, e.g. its center of mass, inertia tensor, and mass.
Since we do not use any dynamic properties here, the interested reader is referred to aforementioned publications for a full discussion of the dynamic model.
The MMM framework is designed to not only work with human subjects but also with objects that are part of a scene.
While scenes that contain objects are currently not part of our dataset, it is still important to note that such data could be used in the future.

The implementation of the MMM framework is open source\footnote{\url{https://gitlab.com/groups/mastermotormap}} and provides ready-to-use converters for our motion capture setup that implement the aforementioned least-squares optimization.
It also specifies an XML-based data format, which we use in our dataset to store motion recordings as represented using the MMM reference model.
The framework includes a range of tools, amongst other 3D visualizer that can be used to inspect MMM motion data.
Terlemez~et~al.~\cite{DBLP:conf/humanoids/TerlemezUMDVA14} discuss the data format and the available tools in depth.

\section{Natural Language Annotations}
\label{sec:natural-language}
This section is concerned with the acquisition of motion descriptions using natural language annotations.
While we resorted to existing data sources for the acquisition of motion data, we had to collect the annotations ourselves.
For this purpose, we created a web-based tool specifically designed for this task called the \emph{Motion Annotation Tool}.\footnote{\url{https://motion-annotation.humanoids.kit.edu}}
We describe the user interface of the tool, cover gamification approaches that we used to motivate annotators, and finally describe a novel approach that we used to decide which motion to present next to the user for annotation.
\Revision{The source code of the Motion Annotation Tool is openly available.}\footnote{\RevisionOnly{added URL}\url{https://gitlab.com/h2t/MotionAnnotation}}

\begin{figure*}[t]
  \centering
  \includegraphics[width=0.8\textwidth]{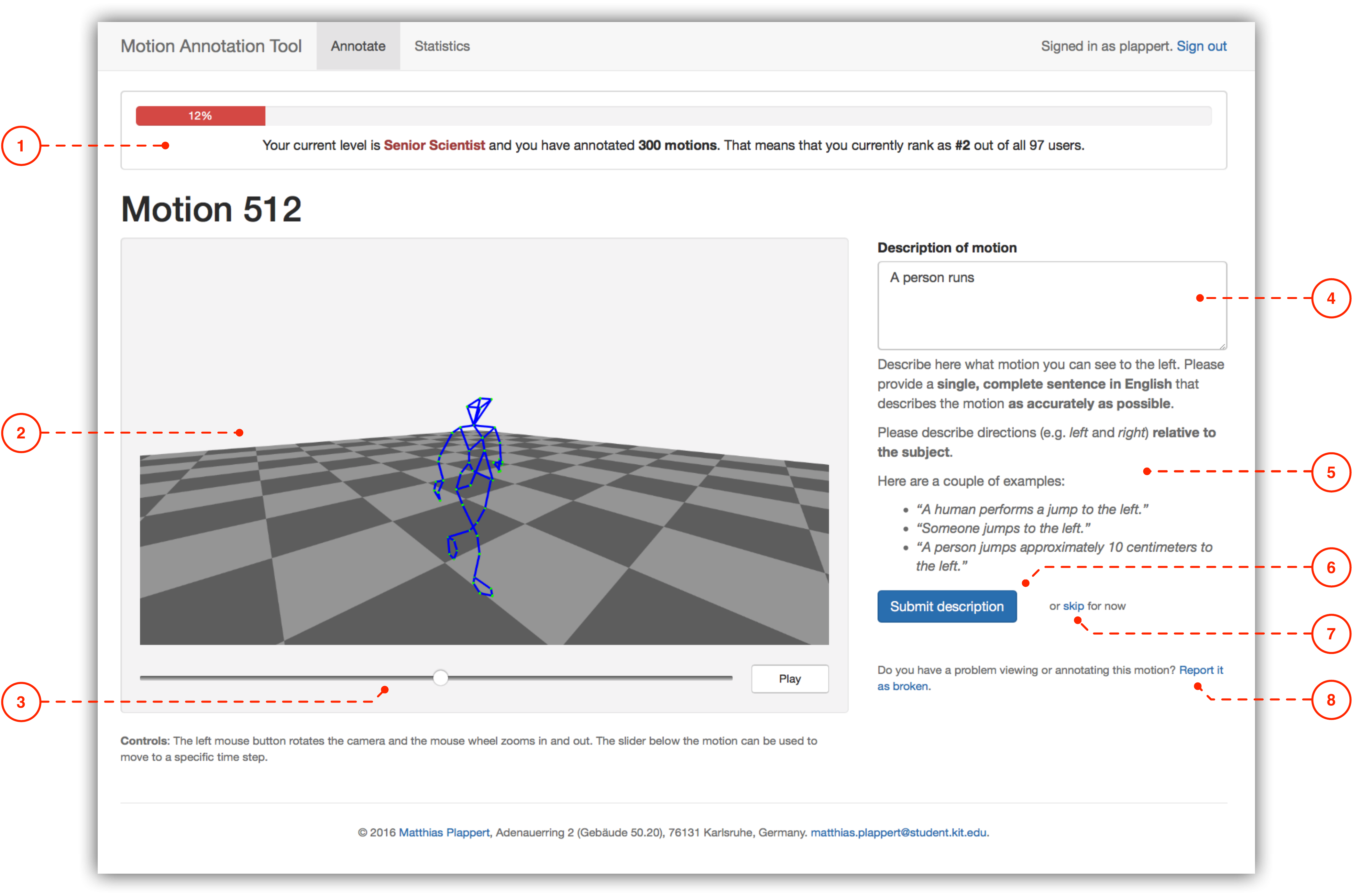}
  \caption{A screenshot of the Motion Annotation Tool during annotation. Visible are the annotator's progress (1), the interactive visualizer (2), the playback controls (3), the input field (4), the annotation instructions (5), the submit button (6), the skip button (7), and the report-a-problem button (8).}
  \label{fig:screenshot_mat}
\end{figure*}

\subsection{User Interface}
The Motion Annotation Tool was designed to be as easily accessible as possible to a wide audience.
We therefore decided to implement the tool using web-based technologies so that no additional software would need to be installed on a volunteer's computer.
We also ensured that the tool works equally well on modern portable devices like smartphones and tablets so that volunteers could annotate while on the go, e.g. during a commute from or to work.

Another interesting problem that we had to solve was how the motion would be visualized to the human annotator.
Consider, for example, a wiping motion and another motion where the subject runs.
In the first case, the subject is stationary and the interesting part of the motion is the movement of the hand.
In contrast to this, in the second case, the motion is highly non-stationary where the subject travels a large distance.
Additionally, the entire body is part of the motion and a different focus of the annotator is likely necessary.
This is further complicated by the fact that some motions are performed using either the left or right hand, making it necessary to show the relevant parts of the body.
We therefore decided early on that the annotation visualization must be interactive instead of using a rendered video of the motion from a fixed perspective.
We used the \emph{Web Graphics Library} (WebGL)\footnote{\url{http://webgl.org}} and the \emph{three.js} framework,\footnote{\url{http://threejs.org}} which builds on WebGL, to implement an interactive 3D~visualizer directly in the user's web browser without the need to install additional software.
This allows the user to select an appropriate perspective that help him or her with the annotation process by rotating and zooming the virtual camera freely.
\Revision{We use raw point cloud data and visually connect points to produce a skeleton-like visualization.
We decided not to use MMM model for visualization since this would require downloading a relatively large COLLADA model and would also significantly increase the computational cost during visualization, which is especially problematic on mobile devices.}
Below the visualizer, a pause/play button and a slider, which allows to skim through the motion quickly, can be used to control the playback.
A screenshot of the Motion Annotation Tool's user interface elements during annotation is depicted in \Cref{fig:screenshot_mat}.

Besides this main user interface element, the interactive visualizer, the user is presented with a text field where he or she can enter the annotation in natural language.
We include guidance to the annotators by asking them to provide a description in the form of a single and complete sentence in English that describes the motion as accurately as possible.
We also provide some exemplary sentences to the user, which are not related to the current motion but simply illustrate the required format.
Basic validation is applied to each annotation before saving it to the database.
This is done to avoid cluttering the dataset with erroneous annotations, both unintentionally (e.g. spelling mistakes) and intentionally (e.g. typing in nonsense) created ones, and ensures a minimum quality standard.
The validation uses some simple heuristics like the total number of words, the percentage of correctly spelled words (as determined by an open-source dictionary of the American English language), and the occurring punctuation to decide if a user's input is consistent with the aforementioned requirements.
The validation is designed to be conservative since too strict requirements may exclude valid annotations.
Once the user has finalized an annotation, it can be submitted.
Alternatively, the user interface also features buttons to either skip a motion, which allows him or her to come back to it later, or to report a problem with the motion data itself.

\subsection{Gamification}
Since our efforts of collecting a large-scale dataset are completely dependent on the willingness of the voluntary annotators, we thought about how we could motivate people to participate.
Besides an easy-to-use user interface, as discussed previously, we utilize \emph{gamification} to make the annotation process more interesting and fun.
Deterding~et~al.~\cite{DBLP:conf/mindtrek/DeterdingDKN11} define gamification as ``the use of game design elements in non-game contexts''.
In the Motion Annotation Tool, we use several such game design elements.

Originally, the Motion Annotation Tool simply presented the next motion after the user submitted an annotation.
In conversations with early users, it became clear that they had a desire to known how many annotations they had already written.
However, simply presenting the user with the number of annotations was often problematic since progress seemed very slow due to the large number of motions that each user could still annotate.
The solution that we chose was the use of a \emph{leveling system}.
A new user starts as a \emph{Novice} and can work his or her way up to \emph{Research Assistant}, \emph{Junior Scientist}, \emph{Senior Scientist} and so on.
Leveling up is easy at first, with only 10 annotations necessary to become a \emph{Research Assistant}.
This means that progress is very apparent initially and we can present a progress bar to the user that fills up quickly (see \Cref{fig:screenshot_mat}).
However, as the user progresses, leveling up becomes harder making it more of a personal challenge for the user to reach higher levels.
The leveling system also led to the introduction of a \emph{leader board}.
The leader board lists all users and their respectively achieved highest level sorted by the number of annotations they have submitted in descending order.
This allows an annotator to see how she or he ranks amongst their peers, which was another request by early users.

While hard to quantify, we believe from conversations with annotators as well as from personal experience that this feature was indeed very helpful to keep annotators going and to make this rather dull process more playful.
It was very apparent to the authors that a sense of personal progress as well as comparison with other annotators was a strong motivator for most participants and ultimately a big factor in the success of the data acquisition.
We explicitly decided to share these design decision in this article, since we believe that this is an important factor in data acquisition, which is often overlooked.

\subsection{Motion Selection}
\label{subsec:motion-sampling}
While we have already discussed how the annotation itself works, an important question has not been answered yet: How is the next motion for annotation selected?
In our work, we used a mixture of two different selection strategies.

The first approach is very much straightforward and works by randomly selecting any motion that has not been annotated so far, with all motions having equal probability of being selected.
We use this approach to initialize the annotation process.
However, as soon as each recorded motion has been annotated once, it becomes unclear which motion should be selected for further annotation.
An obvious approach is to simply sample from the set of motions that have the fewest annotations.
However, this approach has two problems: First, the motion data that we use contains a lot of recordings of bipedal locomotion (e.g. walking and running) but much less recordings of other motions (e.g. dancing and kicking).
This is due to the nature of our research activities, which involved the extensive study of such motions.
Second, we noticed that some annotations stood out due to a different interpretation of the motion by the annotator (which is valid) and also due to low-quality annotations with spelling and grammatical errors (which is unwanted).
We therefore decided to use the annotation data that we already collected to estimate good candidates for further annotation.

To do so, we use the \emph{perplexity}~\cite{DBLP:journals/coling/BrownPdLM92} of an annotation, which is defined as:
\begin{equation}
    \ppl_i := P(\vec{a}_i)^{-1 / |\vec{a}_i|},
    \label{eq:ppl}
\end{equation}
where $\vec{a}_i$ is the concatenation of words of the $i$-th annotation, $|\vec{a}_i|$ denotes the number of words in $\vec{a}_i$ and $P(\vec{a}_i)$ denotes the probability of $\vec{a}_i$ under some statistical language model like an $n$-gram model~\cite{DBLP:journals/coling/BrownPdLM92}.
Briefly speaking, an $n$-gram language model is based on the simplifying assumption that the probability of a word in a sentence only depends on the $n-1$~words before it.
This makes it feasible to compute the word probabilities for small $n$ by simply counting all occurrences in the text corpus.
More concretely, we use a $4$-gram language model to predict the probability $P(\vec{a}_i)$, which is trained on the set of all $N$ annotations $A = \{\vec{a}_1, \ldots, \vec{a}_N\}$.
The perplexity can be thought of as a measure of ``surprise'' under a given model.
If the text of an annotation can be predicted with probability $P(\vec{a}_i) = 1$, it follows that $\ppl_i = 1$.
In contrast, if the probability becomes smaller than $1$ because the model is less confident in predicting the text, the perplexity increases.

We use this property to prefer motions with higher perplexity as candidates for further annotation.
We define the perplexity of a the $j$-th motion simply as the mean over the perplexity of all its annotations:
\begin{equation}
    \mppl_j = \frac{1}{|A_j|} \sum_{\ppl_i \in A_j}{\ppl_i},
    \label{eq:mppl}
\end{equation}
where $A_j$ is the set that contains all annotations that are associated with the $j$-th motion and $|A_j|$ denotes the cardinality of the set.
Notice that each motion has at least a single annotation due to the fact that we used the first selection approach to initialize, hence $|A_j| \geq 1$.
The selection is then realized by sampling from a discrete probability distribution over all $M$ motions:
\begin{equation}
    P(j) = \frac{\mppl_j}{\sum_{k=1}^M{\mppl_k}},
    \label{eq:proba}
\end{equation}
where $P(j)$ denotes the probability of selecting the $j$-th motion.
From \Cref{eq:proba} and $\forall~j~\in~\{1,~\ldots,~M\}:~\mppl_j~\geq~1$ (compare \Cref{eq:ppl,eq:mppl}), it follows directly that $P(\cdot)$ fulfills all three properties of a discrete probability distribution (\emph{Kolmogorov} axioms).
Once a motion has been annotated, we temporarily exclude it from the sampling process until the perplexities have been re-computed, which happens periodically at every hour.

Motions that are over-represented in the dataset will likely have lower perplexity to begin with since their type of motion have already accumulated more annotations.
In contrast, motions that have ``surprising'' annotations either because they are under-represented or because the annotations contain errors will have high perplexity, leading to their selection.
Due to the fact that we re-compute the perplexity every hour, the selection process continuously adapts itself.
We show that the perplexity-based sampling approach is indeed effective in \Cref{sec:perplexity}.

\section{Dataset}
\label{sec:dataset}
While previous sections discussed human motion and natural language annotations individually, we now bring our attention to the resulting dataset, which combines both.
We first provide an overview of its structure, some statistics on the data that it contains, and finally compare its properties to other datasets.

\subsection{Structure}
The dataset is organized as follows.
For each entry, the dataset contains four files: The raw motion data as produced by the capture system, the converted motion using the MMM reference model, the annotations in natural language, and additional metadata.
\Cref{fig:dataset} depicts five exemplary motion and their respective annotations from our dataset.

The motion data itself is available in two different formats.
The format of the raw motion data varies depending on the source of the motion.
Currently, the dataset only contains motion data that was recorded using an optical marker-based motion capture system, which is stored in C3D files.
\Revision{Since this is an industry standard, a wide variety of tools and libraries (e.g. visualizers or file parser) exist for this data format. 
Furthermore, our Motion Annotation Tool uses this data during visualization as well.}
However, since the dataset combines data from different data sources with different capture setups, the marker set varies across recordings.
Furthermore, we intend to include data from additional sources in the future, which might use a different capture modality and data format altogether (e.g. IMU-based motion capture).
We therefore include XML files which contain the MMM representation for each motion.
Like previously discussed, this frees researchers from having to work with a variety of different formats and tools and provides a unified representation that is independent of the marker placement and even capture system (compare \Cref{sec:motion}).
We encourage researchers to use the MMM representation instead of the raw data since this allows us to gradually expand the dataset in the future by adding data from additional data sources while still maintaining a compatible representation with previous versions, making it trivial for researchers to work with updated versions of the dataset.
As mentioned before, the MMM framework is open source and well-documented and also readily contains cross-platform tools for common tasks, e.g. a tool to play back recorded motions.
Providing this unifying representation for data aggregated from multiple sources is a key property of our dataset.

\begin{table*}[t]
  \centering
  \begin{tabularx}{\linewidth}{p{0.22\linewidth}R{0.17\linewidth}p{0.03\linewidth} | p{0.03\linewidth} p{0.22\linewidth}R{0.17\linewidth}}
    \toprule
    \multicolumn{2}{c}{\textsc{Human Motion}} & & & \multicolumn{2}{c}{\textsc{Natural Language}} \\
    \midrule
    \textbf{\# Recordings} & $3911$ & & & \textbf{\# Annotations} & $6278$ \\
    \textbf{\# Subjects} & $111$ & & & \textbf{\# Annotators} & $110$ \\
    \textbf{Total Duration} & $11.23$h & & & \textbf{\# Words} & $52\,903$ \\
    \textbf{Mean Duration} & $10.33~\pm~13.38$s & & & \textbf{Vocabulary} & $1623$ \\
  \bottomrule
  \end{tabularx}
  \caption{\RevisionOnly{update data}Overview of the dataset content.}
  \label{table:dataset_overview}
\end{table*}

The second important part of the dataset are the annotations in natural language.
Each entry is associated with a set of such annotations, making it a one-to-many relationship.
We provide this data in a very simple, JSON-based format:
For each entry, the respective file contains all associated annotations as a simple array of strings.
We chose to use JSON files for this purpose since they are trivial to parse and human-readable.

Our dataset also contains metadata for each entry, where each file contains information that links the entry to the Motion Annotation Tool by providing the respective ID for the entry itself and the IDs of the associated annotations.
The IDs are unique, even across different data sources, and are guaranteed to be permanent.
This makes them especially suitable to reference specific entries or a set of entries, which is useful in a machine learning context, where the dataset is usually split into a training, validation and test subsets.
We consciously decided to avoid defining a split ourselves since some researchers may only work with a suitable subset of the data.
\Revision{We also make available the perplexity scores of each annotation.
This allows researchers to prune the data (e.g. removal of annotations with unusually high perplexity) depending on their individual needs.}
The metadata also clearly states the institution that recorded the motion data and contains the necessary IDs to look up the motion in the respective source database.
This allows researchers to retrieve additional information that could be of interest but is not part of our dataset.
For example, the previously mentioned KIT database contains information on the subject like height, weight, and gender.
It also provides the so-called \emph{Motion Description Tree}, which is a categorization system that assigns each motion one or multiple hierarchically organized tags (e.g. ``walk'' and ``left'') as well as additional recordings like videos and sometimes force sensor data~\cite{DBLP:conf/icar/ManderyTDVA15}.
All of this data can be easily accessed using the provided metadata and the available API.
Motions that originated from the CMU database can be looked up in similar fashion and we intend to include appropriate metadata for new data sources in the future as well.
\Revision{If, for example, a new data source that includes camera calibration is added, the reference to the original dataset in our metadata makes it possible to retrieve and use this calibration data as well.}

We provide the dataset as a compressed ZIP archive, which can be directly downloaded from the Motion Annotation Tool.\footnote{\RevisionOnly{changed URL}\url{https://motion-annotation.humanoids.kit.edu/dataset}}
Beside the latest release, we also provide an archive of older versions of the dataset.
Each version is uniquely identified by its release date, making it possible for researchers to reference and obtain a specific version.
The download page also contains a detailed and up-to-date technical documentation on the structure and data formats of the individual files.

\subsection{Content}
\Revision{As of October 10, 2016, our dataset contains $3911$~motions, of which $2094$~were recorded by the Karlsruhe Institute of Technology (KIT), $1711$ by the Carnegie Mellon University (CMU) and $106$~by the Eberhard Karls Universität Tübingen (EKUT).}
The motions that were recorded by EKUT are part of a joint project and therefore part of the KIT database.
\Revision{All motions combined have a length of $11.23$~hours, with the average motion having a length of $10.33~\pm~13.38$~seconds; the longest motion has a duration of $215.10$~seconds.}
The dataset contains a wide variety of different motions from categories like gesticulation (e.g. pointing and waving), locomotion (e.g. walking and crawling), manipulation (e.g. throwing and wiping), and sports (e.g. martial arts and tennis).
Many motions also available in variations where factors like speed (e.g. walking slow, medium and fast), body part (e.g. using the left or right hand for throwing), direction (e.g. jumping to the left or right), and number of repetition for periodic movements (e.g. wiping the table once or five times) differ across recordings.
The motions were performed by $111$~different subjects, of which $11$~are female and $21$~are male.
The average subject is $27.19~\pm~7.63$~years old, weighs $70.28~\pm~11.15$~kilograms and is $1.76~\pm~0.08$~meters tall.
Data concerning the subjects is only available for recordings from the KIT database, subjects from the CMU database were excluded from these statistics.

\Revision{The dataset contains a total of $6278$~sentences in English.
All sentences combined consist of a total of $52\,903$~words, while the average sentence consists of $8.43~\pm~4.34$~words.
The total vocabulary size, that is the number of unique words disregarding capitalization and punctuation, is $1623$.
The annotations were written by $110$~volunteers, with the an average of $57.43~\pm~84.63$~annotations per volunteer.
Most annotations were written by volunteers located in Germany~($5270$), followed by France~($78$), Australia~($58$), Austria~($39$), the United States~($24$), United Kingdom ($11$), Singapore ($5$), and Japan ($5$).
The origin of the remaining $788$~annotations is unknown since the location can only be estimated from IP addresses.
We also analyzed the the configured language preference of the volunteers (as indicated by their \texttt{Accept-Language} HTTP header): $4009$~annotations were written by people who had their language preference set to German, $1826$~to English, $101$~to French, $50$~to Chinese, and $38$~to Polish.
The language preference of the creators of the remaining $254$~annotations could not be determined.}

\Cref{table:dataset_overview} summarizes the most important statistics for both modalities.
It should be clear that our dataset contains a large number of human motions and annotations thereof in natural language.
Furthermore, our dataset is diverse in the sense that it contains a wide variety of different motions that were performed by a large number of different subjects.
Similarly, the annotations were written by a large number of volunteers, which ensures that the dataset contains a diverse set of annotations.
Our resulting dataset has a total size of $8.08$~GiB.
The compressed ZIP archive, which we make available online, is $3.88$~GiB large.

\subsection{Comparison with other Datasets}
We have already discussed other datasets that were used by other researchers in \Cref{sec:related-work}.
However, we have yet to compare them with our dataset.

We believe that using an openly available dataset is important in order to conduct transparent research that can be verified by others.
The lack of an open dataset also makes it impossible to compare results since each author uses different data.
We therefore think that an important first difference is that our dataset is completely open and can be used by anyone, making it suitable to serve as a benchmark for new methods and approaches in the future.

Furthermore, our dataset is not only open but also, to our best knowledge, the largest and most diverse by a far.
\Revision{As previously discussed, our dataset contains $3911$~motion from a wide range of different scenarios and $6278$~sentences.}
Additionally, this will only increase with time since our dataset is designed to grow by adding additional motion data from existing or new data sources and a continued annotation effort using our crowd-sourcing approach.
We intend to continuously release updates as new data becomes available so that other researchers benefit as well.

By using the MMM representation for motion data, we provide a unified representation of human motion that abstracts from the concrete motion capture method.
This relieves researchers from the hassle of having to support a wide variety of different file formats and differing marker sets.
The MMM representation therefore makes it trivial for users of our dataset to update to new releases and benefit from more data in the future, even if a new data source was added.
Additionally, researchers can use the freely available tools from the MMM framework to efficiently work with the motion data.

Finally, our dataset is the most thoroughly documented one. We provide an extensive discussion of our methods and the nature and representation of the included data in this work. From an engineering perspective, we also provide a comprehensive technical documentation of the dataset online, which helps users work with the data.

\FloatBarrier

\section{Perplexity Analysis}
\label{sec:perplexity}
As we have already discussed in \Cref{subsec:motion-sampling}, we use the perplexity of a motion in an attempt to mitigate two problems: The fact that some motions are under-represented in the dataset and potentially erroneous annotations.
In this section, we show that our approach is indeed effective at mitigating the two aforementioned problems.
First, we show that perplexity is a useful measure to find unusual annotations.
Next, we show that using perplexity-based motion selection reduces the mean perplexity and corresponding variance.

\begin{table}[t]
\begin{tabularx}{\columnwidth}{rrX}
  \toprule
    &   \textbf{Perplexity} & \textbf{Annotation}                                                                       \\
  \midrule
  \textbf{1} &      2.38 &  a person walks in a circle to the right                                          \\
  \textbf{2} &      2.39 &  a person walks in a circle to the left                                           \\
  \textbf{3} &      2.42 &  a person walks a quarter circle to the right                                     \\
  \textbf{4} &      2.44 &  a person walks a quarter circle to the left                                      \\
  \textbf{5} &      2.47 &  a person walks a quarter circle counter clockwise with 4 steps                   \\
  \textbf{6} &      2.47 &  a person walks forward takes a 180 degree turn to the right and keeps on walking \\
  \textbf{7} &      2.47 &  a person walks forward during that the person is pushed to the left              \\
  \textbf{8} &      2.49 &  a person walks in a quarter circle to the right                                  \\
  \textbf{9} &      2.49 &  a person walks in a quarter circle to the left                                   \\
 \textbf{10} &      2.49 &  a person walks forward takes a 180 degree turn to the left and keeps on walking  \\
\bottomrule
\end{tabularx}
\caption{The ten annotations with the lowest perplexity, in ascending order.}
\label{table:lowest-perplexities}
\end{table}

\begin{table}[b]
\begin{tabularx}{\columnwidth}{rrX}
  \toprule
    &   \textbf{Perplexity} & \textbf{Annotation}                                                                       \\
\midrule
  \textbf{1} &     1160.68  & slow walking motion                                          \\
  \textbf{2} &      981.18 & fast walking motion                                          \\
  \textbf{3} &      648.36 & slowly backing off                                           \\
  \textbf{4} &      571.38 & walking with turn at end                                     \\
  \textbf{5} &      469.69  & this motion is broken                                        \\
  \textbf{6} &      422.22 & he walks straight                                            \\
  \textbf{7} &      339.79 & something is happening to the person during the forward walk \\
  \textbf{8} &      333.23 & person dancing the                                           \\
  \textbf{9} &      309.08 & walk turn and walk back                                      \\
 \textbf{10} &      303.95 & somebody stops suddenly                                      \\
\bottomrule
\end{tabularx}
\caption{The ten annotations with the highest perplexity, in descending order.}
\label{table:highest-perplexities}
\end{table}

\subsection{Annotation Perplexity}
\Cref{table:lowest-perplexities} lists the ten annotations that have the lowest perplexity in our dataset, hence annotations that have high probability.
We list all annotations converted to lowercase and without punctuation since this is the form that we use to compute the perplexity.
Also notice that we list each annotation only once and remove subsequent duplicates from the list.
The results are not surprising: All ten annotations describe walking in various forms.
This was expected since the motion data that we use contains large quantities of locomotion recordings.
Additionally, all ten sentences do not contain any spelling or grammatical errors.

In contrast, \Cref{table:highest-perplexities} lists the ten annotations that have the highest perplexity, with the same preprocessing steps (as described before).
It is immediately apparent that the content of this table is vastly different from \Cref{table:lowest-perplexities}.
The first $4$~sentences are clearly no complete English sentences but instead descriptions in note form.
Another interesting example is sentence number $5$: Here, the annotator noticed that the motion is broken and used the annotation field to communicate this.
However, this is obviously a strange annotation, so the high perplexity is expected and desired.
Annotation number~$8$ appears to be an erroneous annotation where the annotation was presumably submitted prematurely by accident.

\begin{table}[h]
\begin{tabularx}{\columnwidth}{rrX}
  \toprule
    &   \textbf{Perplexity} & \textbf{Annotation}                                                                       \\
  \midrule
  \textbf{1} &       2.85 & a person walks 4 steps forward \\
  \textbf{2} &       5.51 & a person walks forward then turns right by 180 degrees and walks forward again \\
  \textbf{3} &      12.24 & a human walks a very slow 90 degree arc to the right \\
  \textbf{4} &      31.33 & a person walks without a hurry \\
  \textbf{5} &      54.68 & subject walks backwards slightly angled \\
  \textbf{6} &     154.34 & some is reverse walking \\
  \textbf{7} &     309.08 & walk turn and walk back \\
  \textbf{8} &    1160.68 & slow walking motion  \\
\bottomrule
\end{tabularx}
\caption{Different annotations of walking motions, sorted by their perplexities in ascending order.}
\label{table:walk-perplexities}
\end{table}

\begin{figure}[h]
  \centering
  \includegraphics[width=\columnwidth]{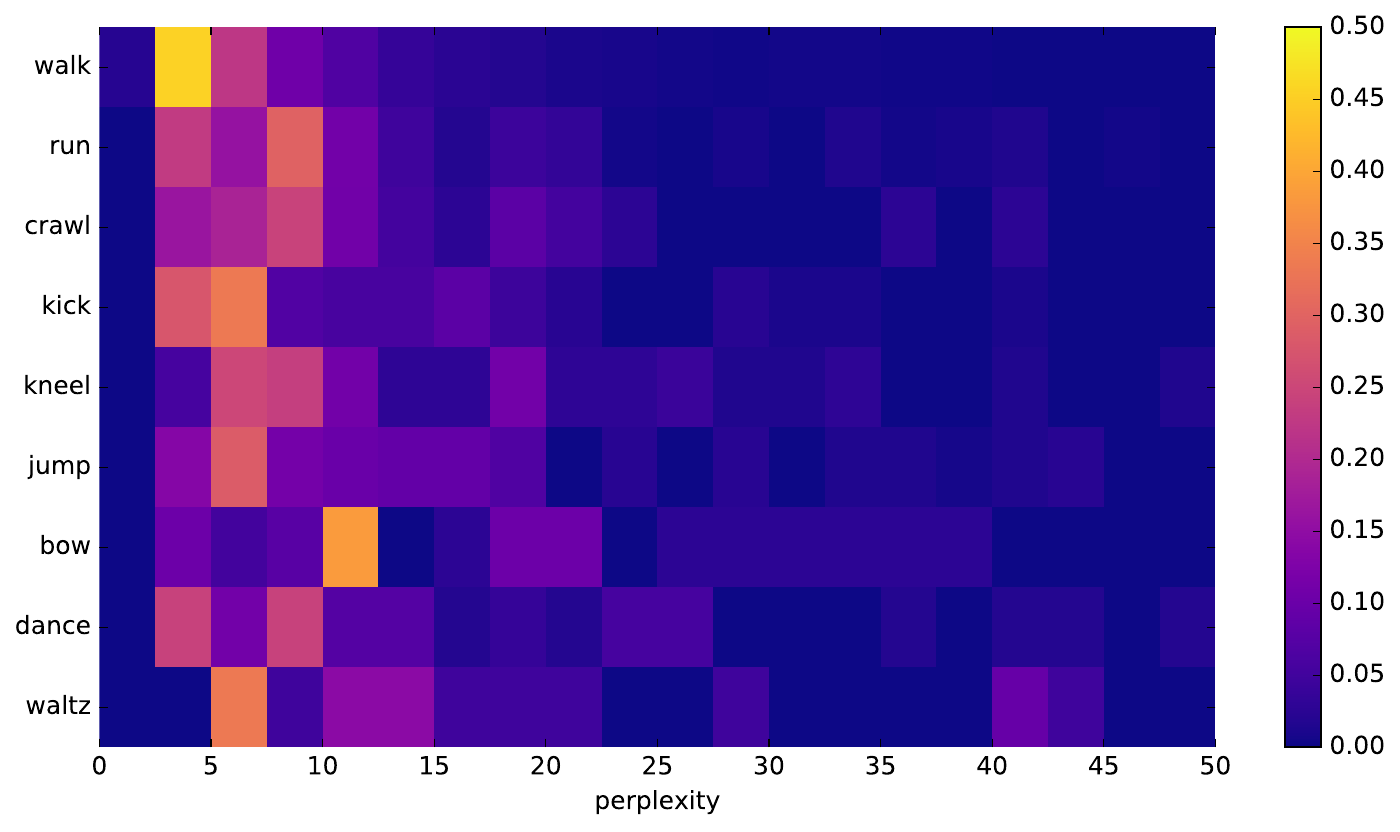}
  \caption{A heatmap that shows in which perplexity range annotations lie that contain the depicted keyword. Bright areas (yellow) correspond to a high occurrence, whereas a dark areas (violet) correspond to no occurrence at all.}
  \label{fig:perplexity_words}
\end{figure}

This pattern of correlation between perplexity score and the subjective quality of the annotation continuous beyond the few samples listed.
\Cref{table:walk-perplexities} illustrates this by listing annotations with increasing perplexity that all describe walking motions.
Notice that the annotations look decent at first but worsen with increasing perplexity.
Similarly, \Cref{fig:perplexity_words} depicts the relative occurrence of a keyword in an annotation (e.g. ``walk'') with respect to the perplexity of the annotation.
This creates a heatmap that, similar to a histogram, shows in which perplexity range the annotations that contain a given keyword lie.
Since some keywords occur much more frequently than others, we normalize by dividing the number of occurrences in each perplexity range by the total number of occurrences.
It is immediately apparent that annotations that describe walking motion mostly have low perplexity.
On the other hand, annotations that contain the keyword ``dance'' and especially ``waltz'' have much higher perplexity.
This again corresponds to the fact that the motion data contains much less motion recordings of people dancing than walking.

Our analysis clearly demonstrates that the proposed perplexity score serves as a useful measure to identify motion candidates for future annotation by the user.

\subsection{Perplexity-based Motion Selection}
\begin{figure}[h]
  \centering
  \includegraphics[width=\columnwidth]{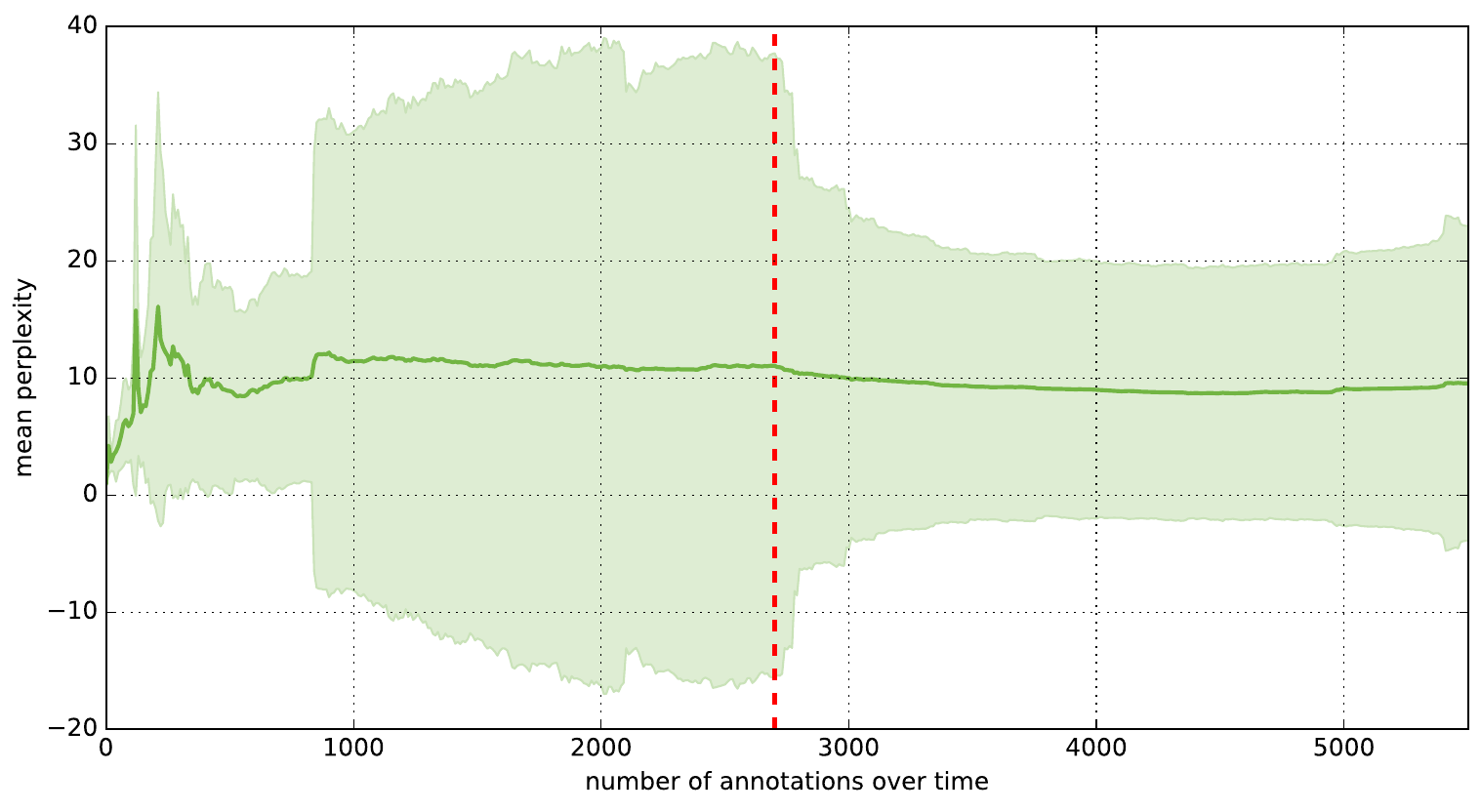}
  \caption{The mean motion perplexity (line) and standard deviation (shaded area) plotted over the accumulated number of annotations over time. The dashed vertical line (red) marks the point in time at which we switched from random-based selection to perplexity-based selection.}
  \label{fig:perplexity}
\end{figure}

The Motion Annotation Tool uses two different selection strategies: \emph{Random-based selection}, in which the next motion for annotation is selected uniformly from the pool of motions with the fewest annotations and \emph{perplexity-based selection}, in which the probability of selecting a motion is proportional to its mean perplexity (compare \Cref{subsec:motion-sampling}).
We initially used random-based selection and switched to perplexity-based selection on April 25 2016.
It is important to note that the system used random-based selection for a while even though all motions were already annotated (at this point in time, the dataset contained only $2097$~motions since the CMU data had not been imported yet).
This makes it possible to directly compare both approaches in a similar setting.
\Cref{fig:perplexity} depicts the mean perplexity over all motions that have annotations and corresponding standard deviation over time.

In \Cref{fig:perplexity}, three different phases can be identified: In the first phase, the mean perplexity is rather noisy.
This is simply due to the fact that not many annotations have been collected at this point, which means that a single annotation can heavily influence the mean.
This first phase is relatively uninteresting, since not enough data is yet available.

The mean perplexity stabilizes around $1000$~annotations, which leads to the second phase.
However, notice the high standard deviation of the motion perplexities.
This is due to the previously described problem:
Since locomotion dominates the motion data, we have already collected many annotations for this type of motion, which, in turn, decreases their perplexities.
In comparison, however, very few annotations exist for other types of motion, hence they have high perplexity.
This large difference in perplexities results in the high variance visible in \Cref{fig:perplexity}.
Since random-based selection does not differentiate between different motions except for the number of collected annotations per motion, the probability of selecting a motion of type locomotion is much higher than that for any other type of motion.
This, in turn, only worsens the situation and explains the continuously growing standard deviation.

The third phase begins as we switch from random-based selection to perplexity-based selection, which is indicated by the dashed vertical red line.
The effect of this change is striking: The standard deviation of the motion perplexities decreases significantly as more annotation data is collected.
Similarly, the mean motion perplexity decreases noticeably as well.
It is also interesting that the effect of the change is rapid at first and slows down over time.
This observation makes perfect sense:  Since we select motions with a probability proportional to their perplexity, we selectively collect data for motions with high perplexity.
At first, a few motions with very high perplexity exist, either because they are under-represented in the dataset or because their annotations are erroneous.
The probability of selecting these for further annotation is very high, which in turn leads to a significant drop in perplexity.
Conversely, the probability of selecting motions that already have low perplexity is very low, which means that no new annotations for these motions are collected.
As a result, the motion perplexities decrease rapidly at first, slowing down as they approach the lower bound.
This also reduces the variance, since the upper bound of the perplexities approach the lower bound.
Also notice that perplexity-based selection not only decreases the mean perplexity and standard deviation, but also appears to act as a stabilizer.
This can be seen by comparing the mean perplexity before and after the switch to perplexity-based selection.
When using perplexity-based selection, the mean and standard deviation are smooth whereas random-based selection results in larger changes and thus a noisy curve.

The results clearly demonstrate the effectiveness of our perplexity-based selection approach.

\section{Conclusion}
\label{sec:conclusion}
The incorporation of human motion and natural language is an important area of research that has many applications, especially in robotics.
There have been years of research in this area to improve our understanding of how to link motion and language models.
However, no standardized and open dataset exists that enables researchers to evaluate and compare their approaches.

In this work, we presented the \emph{KIT Motion-Language Dataset}, a large, open and extensible dataset for the linking human motion and natural language.
Our dataset is freely available and significantly larger and more diverse than previously used datasets.
It is also highly extensible, which we hope will lead to a continuous growth in size.
We utilized the unifying properties of the Master Motor Map framework to achieve a motion representation that is independent of the concrete motion capture system and thus independent of the data source itself.
Additionally, we thoroughly documented all aspects of the dataset, from the data acquisition process to the structure and contents of the downloadable dataset.
\Revision{As of October 10, 2016, the dataset contains $3911$~motions and $6278$~annotations thereof in natural language}.
We believe that the properties of our dataset make it an excellent candidate for systematic benchmarking in this research area and hope that other authors will adopt it in their work.
The latest version as well as archived versions of the dataset are available online: \url{https://motion-annotation.humanoids.kit.edu/dataset}

We also presented a novel way to decide which motion to present to the user during annotation: perplexity-based motion selection, which is used for the crowd-sourced collection of annotations.
We show that the perplexity is a suitable metric to detect annotations that are either under-represented or controversial.
We further show that the perplexity-based selection approach helps decrease the variance in perplexity and acts in a stabilizing manner.

However, there is still room to improve the dataset in future work.
First, we intend to add more motion data from the wide variety of available data sources.
Second, the Motion Annotation Tool currently only supports motions performed by a single subject and without any objects.
Especially the inclusion of objects is an important next step since they provide additional context, e.g. for manipulation tasks.
Third, we intend to actively develop the dataset and its contents to incorporate ideas and feedback which we will hopefully receive from users of our dataset.
\Revision{Fourth, defining a measure for motion data similar to the annotations' perplexity is an interesting problem.
This would allow the annotation system to decide which motion should be annotated next if not all motions have been annotated yet or if new motion data is added.
A possible approach could be to use unsupervised clustering using Hidden Markov Models (HMMs)~\cite{DBLP:conf/ro-man/KulicTN07, DBLP:journals/ijrr/KulicTN08}.
For an unannotated motion, the distance to existing clusters could be used to decide how to prioritize the motion for annotation.
Furthermore, this approach could also be used to find interesting structure and clusters in the data that could eventually become part of the dataset.
}

\section*{Acknowledgments}
We would like to thank the numerous volunteers that helped make this dataset possible by providing the annotations in natural language.
In particular, we would like to thank the staff members and students at the H\textsuperscript{2}T lab and students of the ``Robotik II'' lecture during the summer term of 2016 at the Karlsruhe Institute of Technology.
While we cannot list each annotator, we would like to acknowledge the five people that contributed the most annotations (excluding the authors): Max Hertel, uaejd (anonymous), Vera Schumacher, Laura Tessin, and Inna Belyantseva.
A complete and up-to-date list of annotators is available online.
The research leading to these result has received funding from the H\textsuperscript{2}T lab, the German Research Foundation (DFG) in the SPP 1527 (Autonomous learning) and the European Union's Horizon 2020 Research and Innovation Programme and Seventh Framework Programme under grant agreements no. 611832 (WALK-MAN), no. 611909 (KoroiBot), no. 643950 (SecondHands), and no. 643666 (I-Support).
Part of the motion data used in this project was obtained from the CMU MoCap Database (mocap.cs.cmu.edu), which was created with funding from NSF EIA-0196217.

\bibliographystyle{unsrt}
\bibliography{mybibfile}

\end{document}